%% file: main.tex
\documentclass[a4paper,conference]{IEEEtran}

\usepackage{amsmath}
\usepackage{booktabs}
\usepackage{cite}
\usepackage{graphicx}
\usepackage{url}
\usepackage{xcolor}

\newcommand{\system}{CoreSense}
\newcommand{\gate}{Conflict-Aware Belief Gate}

\title{CoreSense: Traceable Failure Recall and Conflict-Aware\\
Belief Gating for Auditable Robot Decisions}

\author{\IEEEauthorblockN{Zoe Li}
\IEEEauthorblockA{Independent Researcher, Seattle, WA, USA\\
zoeli4@siggraph.org}}

\begin{document}
\maketitle

\begin{abstract}
Robots can recall prior failures without knowing whether recalled evidence remains valid, conflicts with current observations, or is sufficient to guide a decision. We present \system, a robot-system integration architecture that combines traceable episodic evidence with a conflict-aware belief gate and bounded, auditable recommendations. The gate checks scope, provenance, time, contradiction, and support before it permits \textsc{Proceed}, requests re-observation, abstains, or escalates. Evaluation follows three complementary layers without commanding a physical robot: offline public real-robot data, a frozen signal-level simulation, and a live cloud deployment path. On CableTrace-120 and BotFails-200, belief gating reduces protocol-defined unsafe proceeds from 20\% and 40\% to 0\%. A disjointly calibrated raw-video policy also reaches 0\% unsafe proceed, but overblocks every nominal episode. On public data, a ViFailback--BotFails visual detector reaches 0.778 AUROC yet remains all-blocking, whereas cycle-disjoint UR3 telemetry for protective stops yields 0\% unsafe proceed, 36.1\% overblocking, and 61.9\% coverage; grip-loss transfer remains a negative result. Controlled physical corroboration yields 3.3\%, 0\%, and 42.0\%, while conflict-aware fusion yields 4.7\%, 0\%, and 42.8\%. Finally, 20/20 cloud recalls validate a CockroachDB Cloud--Amazon Bedrock deployment path. The evidence supports an auditable integration pattern, not autonomous recovery or certified safety.
\end{abstract}


\section{Introduction}
Failure-aware robot systems increasingly combine perception, learned models, retrieval, and language-conditioned reasoning. Recent work improves failure recognition and explanation with vision-language models~\cite{duan2025aha}, and retrieves failure exemplars to support online detection and causal reasoning~\cite{ying2026robofailring}. These advances make remembered experience useful, but retrieval alone does not answer a system-level question: \emph{when should recalled evidence be allowed to influence action?}

An episode can be similar yet stale, recorded for another object or task, contradicted by current sensors, or supported only by a derived model claim. Returning the nearest episode or producing a fluent recovery suggestion can therefore create a false sense of authorization. A deployable integration layer must preserve source boundaries, distinguish observation from inference, expose unresolved conflicts, and bound the output when evidence is insufficient.

We introduce \system, which integrates structured episodic memory with a temporal, provenance-aware \gate{}. The architecture is deliberately not a robot controller. It returns one of four bounded decisions---\textsc{Proceed}, \textsc{Reobserve}, \textsc{Abstain}, or \textsc{Escalate}---together with the evidence and rules that produced it. The main contributions are:
\begin{itemize}
    \item an integration boundary that separates memory retrieval and candidate generation from action authorization;
    \item a traceable evidence representation over claim, source, time, scope, and support, with deterministic conflict-aware gate rules;
    \item a reproducible evaluation across a controlled cable case, fixed public real-robot visual and telemetry splits, and a calibrated signal-level corroboration simulation; and
    \item an explicit portability contract between an offline backend and a CockroachDB plus Amazon Bedrock adapter, without making cloud services part of the algorithmic claim.
\end{itemize}

The present evidence is a functional systems evaluation, not proof of autonomous physical recovery or unconstrained cross-domain generalization. We make this distinction explicit throughout.

\section{Related Work}
\subsection{Robot failure detection and reasoning}
AHA frames failure detection as free-form vision-language reasoning and constructs scalable failure data with FailGen~\cite{duan2025aha}. RoboFailRing combines failure memory with retrieval and grounded language reasoning~\cite{ying2026robofailring}. BotFails provides multimodal nominal and anomalous episodes across ten tasks~\cite{rolland2026botfails}; ViFailback supplies real-world manipulation trajectories with failure diagnosis and correction annotations~\cite{zeng2026vifailback}. Our work does not replace these detectors. It addresses the downstream question of whether heterogeneous evidence is mutually consistent and sufficient to authorize a bounded recommendation.

\subsection{Large robot datasets and transfer}
Open X-Embodiment aggregates heterogeneous robot experience~\cite{openx2024}, DROID emphasizes diverse real-world scenes and tasks~\cite{khazatsky2024droid}, and LIBERO studies lifelong manipulation learning~\cite{liu2023libero}. ARMBench offers an object-centric manipulation benchmark~\cite{gupta2023armbench}; UR3 CobotOps provides public currents, speeds, temperatures, protective stops, and grip-loss labels~\cite{tyrovolas2024ur3}. These resources motivate portable interfaces, but scale does not resolve provenance or temporal conflict. We use small, frozen splits to test integration rules and independent corroboration without claiming physical recovery.

\subsection{Memory as evidence}
Retrieval-augmented systems commonly rank records by semantic similarity. \system{} instead treats retrieved episodes as evidence objects rather than instructions. Similarity may identify a candidate, but authorization depends on source class, temporal validity, task and entity scope, conflicts, and minimum support. This separation is the central systems distinction from a vector-database-only design.

\section{Problem Formulation}
Let a query $q$ contain the current task, entity scope, observations, and time. The memory bank $E$ stores episodes $e_i$ with structured context, outcome, failure mode, recovery, provenance, and timestamps. A retriever returns $R_k(q)\subset E$. A candidate generator proposes a bounded action $a$ and supporting claims. The belief gate evaluates the evidence set $B$ and emits
\begin{equation}
 d = G(q, a, B) \in \{P,R,A,E\},
\end{equation}
where $P$, $R$, $A$, and $E$ denote \textsc{Proceed}, \textsc{Reobserve}, \textsc{Abstain}, and \textsc{Escalate}.

Each claim is represented as
\begin{equation}
 b_j=(c_j,v_j,s_j,t_j,\sigma_j,w_j,\pi_j),
\end{equation}
where $c_j$ is the predicate, $v_j$ its value, $s_j$ the source, $t_j$ time, $\sigma_j$ scope, $w_j$ support weight, and $\pi_j$ provenance pointer. Two active claims conflict when they share a predicate and overlapping scope but assign incompatible values. Source policy distinguishes authoritative current observations from derived claims and historical episodes. The gate never converts similarity alone into permission.

For a query whose protocol target requires withholding action, a protocol-defined unsafe proceed is $\mathbf{1}[d=P]$. For a nominal query, overblocking is $\mathbf{1}[d\neq P]$. We additionally evaluate conflict F1, decision accuracy, audit completeness, and replay consistency.

\section{System Architecture}
\begin{figure*}[t]
\centering
\includegraphics[width=\textwidth]{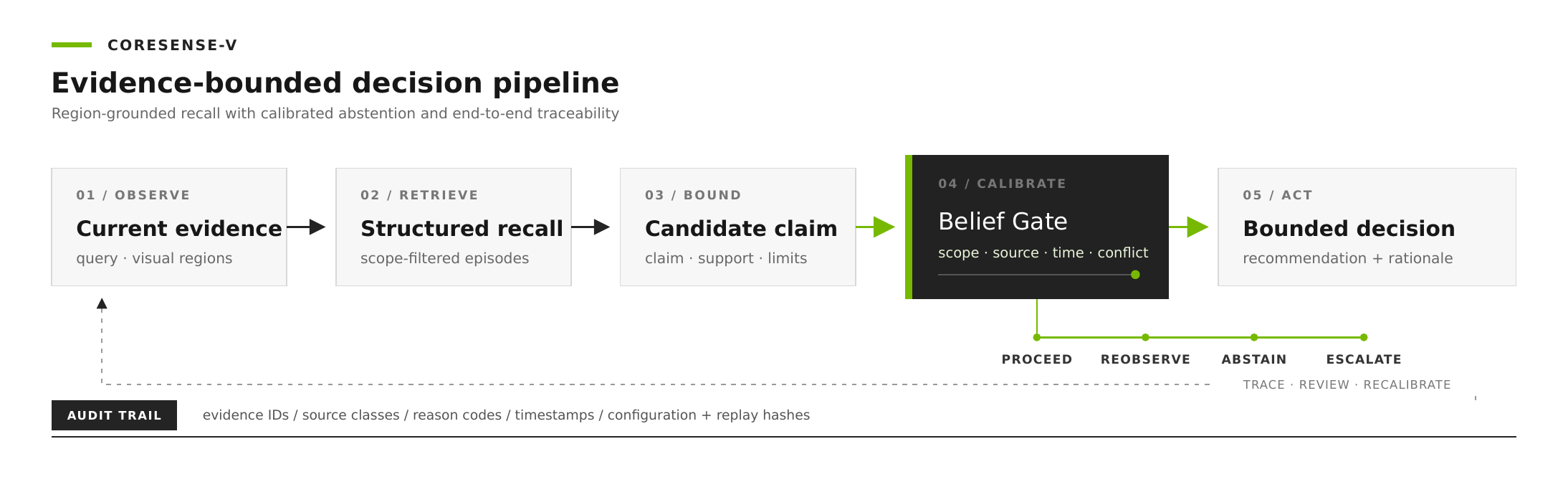}
\caption{CoreSense evidence-bounded decision pipeline. Structured recall produces a bounded candidate, while the belief gate validates scope, source, temporal relevance, and conflicts before authorizing an action. Every decision remains linked to an auditable provenance trail.}
\label{fig:architecture}
\end{figure*}

\subsection{CoreSense episodic memory}
CoreSense stores episodes as typed records rather than free text alone. Retrieval first enforces task and scope constraints and then ranks compatible episodes. The offline implementation uses deterministic lexical and hashed-vector features so that results can be replayed without external services. The production adapter can store the same records in a relational vector store and obtain embeddings from a managed model.

\subsection{Conflict-aware belief gate}
The \gate{} constructs an active claim set from current observations, retrieved episodes, and derived candidate claims. It applies four checks in order: (1) scope compatibility, (2) source precedence, (3) temporal validity, and (4) contradiction/support. Current authoritative observations can resolve older disagreement; a derived claim cannot silently override conflicting direct or historical evidence. Stale or incomplete evidence requests re-observation. Unresolved high-impact conflict abstains or escalates. Only a supported, conflict-free candidate proceeds.

\subsection{Audit contract and autonomy boundary}
Every decision record contains the input and configuration hashes, source episode identifiers, selected claims, reason codes, and final gate state. The system neither synthesizes low-level motion nor sends commands to hardware. This boundary makes a decision inspectable and lets integrators replace storage, embedding, or perception modules without changing gate semantics.

\section{Implementation and Reproducibility}
The reference implementation defines common \texttt{Episode}, \texttt{Query}, and \texttt{Recommendation} schemas. JSONL is the canonical offline representation. All selection and ranking ties are resolved by a fixed SHA-256-based order. Five variants share the same query and output contracts:
\begin{itemize}
    \item M0: stateless candidate generation;
    \item M1: recency and keyword retrieval;
    \item M2: deterministic vector-similarity retrieval;
    \item M3: CoreSense structured retrieval without belief gating; and
    \item M4: CoreSense plus the \gate{}, with the same retrieved set as M3.
\end{itemize}
M3 and M4 therefore isolate the effect of belief gating rather than retrieval quality. Each experiment is executed three times; byte-level hashes of decision files must match.

The cloud adapter maps structured records to CockroachDB \texttt{VECTOR(256)} storage and cosine search~\cite{cockroachvector}, with Amazon Titan Text Embeddings V2 through Bedrock~\cite{awstitan}. The adapter was validated in AWS CloudShell against services co-located in \texttt{us-west-2}. It persisted one synthetic failure record, recalled it in a separate process, and exposed only a bounded \texttt{avoid\_region} intent to a deterministic planner. This portability check is not required for the offline decision logic and does not send robot commands.

\section{Experimental Design}
\subsection{Research questions}
RQ1 asks whether conflict-aware gating reduces protocol-defined unsafe proceeds relative to retrieval or structured memory alone. RQ2 asks whether that reduction causes nominal overblocking. RQ3 asks whether the same rules behave consistently outside the cable workflow. RQ4 asks whether all decision artifacts reproduce exactly. RQ5 asks whether a threshold selected on disjoint raw-video development data transfers to Raw-40. RQ6 asks whether independent detectors trained on public real-robot observations provide selective corroboration. RQ7 asks whether controlled independent physical evidence can improve coverage without exceeding 5\% unsafe proceed.

\subsection{CableTrace-120 v1}
CableTrace-120 is a controlled, sanitized functional benchmark with 60 memory episodes and 60 evaluation queries. It covers six equally sized conditions: nominal operation, missing detection, endpoint-count mismatch, topology conflict, unsupported grasp, and stale contradictory history. The traces are derived from an existing wiring-and-sorting workflow and contain no partner, site, hardware-serial, image, or proprietary process identifiers. They are not 120 independent physical robot trials. Their purpose is to test interface and gate invariants under known conditions.

\subsection{BotFails-200 v1}
BotFails-200 v1 is selected from upstream revision \texttt{3478e49d91e1737eb76dfee2d81bb22617039c13}. The frozen manifest contains 40 development records, 80 memory records, and 80 evaluation queries, balanced across all ten upstream tasks. Evaluation contains four nominal and four native-positive anomalous episodes per task, for 40 of each class. We use label CSVs and episode/task metadata only; raw videos and proprioceptive arrays are not downloaded.

BotFails does not specify a unique correct recovery for every anomaly. We therefore evaluate whether the authorization layer withholds unsupported \textsc{Proceed}, not whether it detects failures from pixels or selects an optimal recovery. To test conflict handling, two anomalous evaluation episodes per task are deterministically assigned a derived false-normal candidate claim, totaling 20 preregistered conflicts. The remaining anomalous queries expose the native anomaly observation. Native labels remain ground truth. This intervention is fixed in the manifest and applied identically to all methods.

\subsection{BotFails-Raw-40 v1 stress test}
We additionally freeze two nominal and two native-positive episodes per task from the BotFails-200 evaluation ordering. For each of the 40 real-robot episodes, we download both RGB views and the state record, then sample each view at fixed fractions $\{0.1,0.3,0.5,0.7,0.9\}$. Amazon Nova Lite receives only task text and JPEG frames through a forced tool schema; labels, split names, source paths, and authorization targets are unavailable. Its cached output becomes a \emph{derived} claim. An M5 ablation processes view 0 first and permits one independent view-1 reobservation. It proceeds only if both claims are nominal and satisfy a confidence threshold; all other outcomes escalate. Invalid tool calls fail closed as zero-confidence uncertainty after three attempts. This is an offline observation replay, not a live robot trial.

For threshold calibration, Raw-Dev-40 v1 contains all 40 records in the frozen BotFails-200 development split: two nominal and two native-positive episodes per task, with zero source-episode overlap with Raw-40. The prompt, frame fractions, model, and retry policy are identical. Before examining its outcomes, we fix the candidate grid $\{0.50,0.55,\ldots,1.00\}$ and select the lowest threshold whose development unsafe-proceed rate is at most 5\%. If none is feasible, the protocol forbids a calibrated Raw-40 run. Raw-40 labels are unavailable to this selection procedure. The resulting lock file, including input and selection hashes, is then consumed once by the held-out evaluation.

\subsection{Independent public-data corroborators}
The visual corroborator uses 240 balanced ViFailback episodes for external training plus six BotFails task groups, with two complete BotFails groups for calibration and two unseen industrial groups for a 12-episode locked test. Four chronological frames are encoded by a frozen MobileNetV3-Small; a logistic head uses sequence mean, variance, and change. It receives no Bedrock output, VLM confidence, belief state, or memory score.

For physical telemetry, UR3 CobotOps is split by complete operation cycle so rows from one cycle never cross train, calibration, and test. Three logistic variants use tool current; all joint and tool currents; or currents, speeds, temperatures, and first differences. Thresholds are calibrated separately for protective-stop and grip-loss targets. Both locked tests are run once. These are offline evaluations of recorded real-robot data, not control trials or calibration for another embodiment.

\subsection{Controlled corroboration simulation}
We freeze a phase-structured signal-level simulator with nominal, empty-grasp, slip, soft-partial-closure, and visual-occlusion-only scenarios. It produces kinematic, actuator-load, contact, and visual-quality \emph{proxy} signals; none are measured current or force. Train, calibration, and locked-test namespaces contain respectively 100, 40, and 100 episodes per scenario, with stronger shift at test. Random-forest corroborators use kinematics, load, or their physical combination. A logistic fusion head consumes out-of-fold physical risk, visual risk, quality, disagreement, and sensor validity. Temperature and the proceed threshold are selected only on calibration data to maximize coverage subject to unsafe proceed $\leq5\%$. The test is executed once. The simulator imports no robot or motion-control interface.

\subsection{Metrics and statistics}
Unsafe-proceed rate is computed over anomalous or unresolved queries; nominal overblock rate is computed over nominal queries. Selective coverage is the fraction permitted to proceed. Detector ranking uses AUROC and AUPRC. Conflict F1 evaluates the preregistered cable and BotFails conflicts; decision accuracy compares the bounded decision with its protocol target. We report Wilson intervals in machine-readable summaries. M3--M4 and M4--M5 use exact paired McNemar tests. Every calibrated threshold is selected on a disjoint split and reported once on its held-out test.

\section{Results}
\begin{table*}[t]
\centering
\caption{Offline results. Protocol-defined unsafe proceed, overblock, and accuracy are percentages; conflict F1 is in $[0,1]$. BotFails conflict results include the fixed derived-claim intervention.}
\label{tab:main}
\input{generated/main_results.tex}
\end{table*}

\begin{table}[t]
\centering
\caption{Paired decision-accuracy comparison of M3 and M4.}
\label{tab:paired}
\resizebox{\columnwidth}{!}{\input{generated/paired_results.tex}}
\end{table}

Table~\ref{tab:main} shows that M0--M3 all permit the ten CableTrace queries whose protocol targets require withholding action, giving a 20\% unsafe-proceed rate under the evaluation definition. M4 gates all ten while preserving proceed decisions for every nominal query. The paired comparison contains ten M4-only corrections and no M3-only corrections ($p=0.00195$).

On BotFails-200, M3 has 40\% protocol-defined unsafe proceeds and 20\% nominal overblocking. M4 resolves the authoritative nominal observations, exposes all fixed conflicts, and withholds every such proceed, yielding 24 M4-only corrections and no M3-only corrections ($p=1.19\times10^{-7}$). These values validate the implemented rules on the frozen intervention protocol. They must not be read as visual failure-detection accuracy because the experiment consumes labels and metadata rather than raw sensor streams.

\begin{table}[t]
\centering
\caption{BotFails-Raw-40 stress test (\%).}
\label{tab:raw}
\begin{tabular}{lrrr}
\toprule
Method & Unsafe $\downarrow$ & Overblock $\downarrow$ & Accuracy $\uparrow$ \\
\midrule
M0 & 40 & 50 & 55 \\
M3 & 20 & 70 & 55 \\
M4 & \textbf{0} & 100 & 50 \\
M5 fixed 0.8 & 30 & 60 & 55 \\
M5 dev-calibrated & \textbf{0} & 100 & 50 \\
\bottomrule
\end{tabular}
\end{table}

On BotFails-Raw-40, the perception adapter predicts 12/20 anomalies and 10/20 nominal episodes correctly (55\% overall); its median and p95 invocation latencies are 3893.9 and 5038.8 ms. M4 reaches 0\% unsafe proceed at 100\% nominal overblocking and 50\% accuracy. With the fixed 0.8 threshold, M5 reobserves 25/40 cases and proceeds on 14: overblocking falls to 12/20 (60\%), but unsafe proceeds rise to 6/20 (30\%); accuracy is 55\% and the paired M4--M5 difference is not significant ($p=0.791$). Median added reobservation latency is 2246.1 ms. Two of 80 view requests returned no valid forced tool call after bounded retries and were recorded as uncertainty. Thresholds 0.6--0.9 produce the same descriptive Raw-40 decisions.

The disjoint Raw-Dev-40 calibration selects 0.95, the lowest candidate satisfying the 5\% safety constraint. At this threshold, development unsafe proceed is 0/20, but nominal overblocking is 20/20 and no episode proceeds. Applying the signed lock once to Raw-40 yields the same boundary: 0/20 unsafe proceeds, 20/20 nominal overblocks, 50\% accuracy, and 38/40 reobservations. The calibration and held-out decision files reproduce byte-for-byte locally. Thus independent calibration prevents evaluation leakage and recovers safety, but it does not make self-reported model confidence a useful authorization signal.

The independent visual detector trained jointly on ViFailback and BotFails improves locked task-disjoint ranking to 0.778 AUROC and 0.869 AUPRC, versus 0.611 AUROC for either source alone. Its 12-episode calibration split nevertheless selects an all-blocking threshold: unsafe proceed is 0\%, overblocking 100\%, and coverage 0\%. The ranking signal is useful, but the selective policy is not.

On cycle-disjoint UR3 CobotOps telemetry, the physical multimodal model reaches 0.994 AUROC and 0.805 AUPRC for protective stops. Its locked point estimates are 0\% unsafe proceed, 36.1\% overblocking, and 61.9\% coverage. Transfer to grip loss fails: AUROC is 0.729, unsafe proceed 16.7\%, overblocking 69.7\%, and coverage 29.6\%. Tool current alone blocks every case for both targets. This failure-specific contrast is retained rather than averaged away.

\begin{figure}[t]
\centering
\includegraphics[width=\columnwidth]{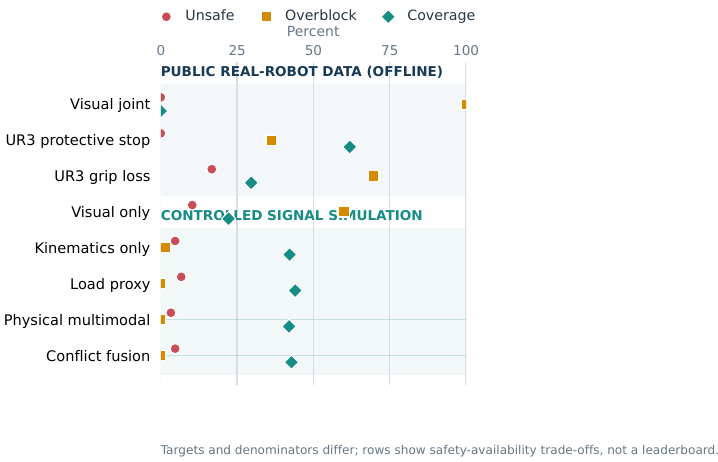}
\caption{Locked-test safety--availability point estimates. Public rows are offline real-robot data; lower rows are controlled proxy simulation. Targets and denominators differ, so the visualization is descriptive rather than a cross-dataset ranking.}
\label{fig:evidence-comparison}
\end{figure}

Figure~\ref{fig:evidence-comparison} also shows the controlled mechanism study. Physical multimodal corroboration yields 3.3\% unsafe proceed, 0\% overblocking, and 42.0\% coverage. Fusion proceeds on all 200 physically safe nominal or visually occluded cases while allowing 14/300 faults: 4.7\% unsafe proceed, 0\% overblocking, and 42.8\% coverage. Its 95\% Wilson unsafe interval is 2.8--7.7\%, so this demonstrates gate mechanics, not a certified bound. Load alone misses the 5\% target.

All three runs in each domain produce identical decision-file hashes. Audit completeness is 100\% by schema validation: every M4 decision contains its query, configuration, evidence, reason, and output identifiers. This establishes deterministic replay of the reference backend, not numerical equivalence with the cloud adapter.

\subsection{Cloud portability check}
After one live record/recall smoke test, we issued 20 sequential read-only recalls from AWS CloudShell. All 20 recalled the same memory with cosine score 0.672078, emitted \texttt{avoid\_region}, and produced the same successful path. Observed end-to-end latency was 1294.5 ms median and 1359.0 ms p95. These values include client startup, Bedrock embedding, network, and CockroachDB retrieval; they are environment-specific descriptive observations, not an availability, cost, or general performance benchmark. The check validates the public CoreSense deployment path, not equivalence of the full cloud and offline belief-gating pipelines.

\section{Discussion}
The results answer a narrow but important systems question. Vector retrieval (M2) is neither necessary nor sufficient for the observed safety behavior: M4 uses structured evidence and an explicit gate, while M2 still proceeds on conflicts. This does not prove that vector databases are generally unnecessary. Semantic retrieval may remain valuable at larger scale; the evidence only shows that authorization semantics cannot be delegated to nearest-neighbor similarity and that a small deterministic backend is sufficient for this benchmark.

The evidence layers answer different questions. CableTrace and BotFails test gate semantics; public RGB and UR3 telemetry test independent signals recorded on real robots; simulation isolates corroboration under controlled shift; the cloud check tests deployability. The UR3 protective-stop result shows that real telemetry can break all-blocking for a specific hazard, while the visual and grip-loss failures prevent a broad detector claim. The proxy simulation supplies the controlled mechanism: independent physical evidence restores coverage under visual occlusion, while load alone remains insufficient.

\section{Limitations and Threats to Validity}
CableTrace is constructed, and BotFails-200 uses metadata and labels. Raw-40 is a small offline replay whose sparse frames may miss transient failures. The independent visual test has only 12 episodes; the UR3 thresholds and current scales are embodiment- and failure-specific, as the grip-loss negative result demonstrates. No experiment measures executed recovery success. The corroboration study is a signal simulator rather than a dynamics engine: proxy load and contact cannot establish real actuator calibration. Cross-dataset rows in Fig.~\ref{fig:evidence-comparison} therefore visualize trade-offs, not statistical superiority.

Rule behavior was designed before final evaluation, but the same author implemented both system and benchmark, creating experimenter bias. Although the threshold is calibrated on a disjoint split, the development set contains only 40 episodes and the confidence scores are model self-reports rather than externally calibrated probabilities. Future work should test selective re-observation policies, add independent annotations and physical recovery trials, and evaluate externally generated conflicts. The cloud check covers one synthetic record and repeated reads; full offline/cloud equivalence, concurrency, cost, outage, retry, and failure-mode tests remain incomplete. Finally, the system offers bounded recommendations only; human and controller behavior after escalation is outside the evaluation.

\section{Conclusion}
\system{} treats remembered failures as traceable evidence rather than executable instructions. Public real-robot data shows both a successful protective-stop corroborator and visual/grip-loss limits; controlled simulation isolates how independent physical evidence restores coverage; a live adapter validates deployment feasibility. Together with the frozen gate evaluations, these complementary layers support an auditable robot-system-integration pattern, not a deployable physical-recovery policy.

\section*{Acknowledgment}
CoreSense originated as an AWS Bedrock hackathon project and used available Amazon Bedrock and CockroachDB Cloud resources for the limited deployment check. The author used OpenAI Codex to assist with implementation scaffolding, experiment scripts, and initial language drafts across the manuscript. The author designed the study, verified the generated code and results, reviewed the final text, and takes responsibility for all claims.

\IEEEtriggeratref{4}

\end{document}

%% file: generated/main_results.tex
\begin{tabular}{llrrrr}
\toprule
Domain & Method & Unsafe $\downarrow$ & Overblock $\downarrow$ & Conflict F1 $\uparrow$ & Accuracy $\uparrow$ \\
\midrule
Cable & Stateless & 20.0 & 0.0 & 0.00 & 83.3 \\
Cable & Recency/keyword & 20.0 & 0.0 & 0.00 & 83.3 \\
Cable & Vector retrieval & 20.0 & 0.0 & 0.00 & 83.3 \\
Cable & CoreSense & 20.0 & 0.0 & 0.00 & 83.3 \\
Cable & CoreSense+Gate & 0.0 & 0.0 & 1.00 & 100.0 \\
\midrule
BotFails & Stateless & 50.0 & 0.0 & 0.00 & 75.0 \\
BotFails & Recency/keyword & 40.0 & 20.0 & 0.00 & 70.0 \\
BotFails & Vector retrieval & 50.0 & 0.0 & 0.00 & 75.0 \\
BotFails & CoreSense & 40.0 & 20.0 & 0.00 & 70.0 \\
BotFails & CoreSense+Gate & 0.0 & 0.0 & 1.00 & 100.0 \\
\bottomrule
\end{tabular}

%% file: generated/paired_results.tex
\begin{tabular}{lrrr}
\toprule
Domain & M3-only correct & M4-only correct & Exact $p$ \\
\midrule
CableTrace-120 & 0 & 10 & 0.00195 \\
BotFails-200 & 0 & 24 & 1.19e-07 \\
\bottomrule
\end{tabular}

%% file: main.bbl
\begin{thebibliography}{99}
\bibitem{duan2025aha}
J. Duan \emph{et al.}, ``AHA: A vision-language-model for detecting and reasoning over failures in robotic manipulation,'' in \emph{Proc. ICLR}, 2025.

\bibitem{ying2026robofailring}
C. Ying, L. Du, Y. Shu, and P. Cheng, ``RoboFailRing: Retrieval-augmented and language grounding failure detection for VLM-enabled robotic manipulation,'' in \emph{Proc. ACL}, 2026, pp. 13188--13202.

\bibitem{rolland2026botfails}
Q. Rolland, F. Mayran de Chamisso, and J.-B. Mouret, ``Failure identification in imitation learning via statistical and semantic filtering,'' in \emph{Proc. IEEE ICRA}, 2026. [Online]. Available: \url{https://huggingface.co/datasets/kantine/BotFails}

\bibitem{zeng2026vifailback}
X. Zeng, X. Zhou, Y. Li, J. Shi, T. Li, L. Chen, L. Ren, and Y.-L. Li, ``Diagnose, correct, and learn from manipulation failures via visual symbols,'' in \emph{Proc. IEEE/CVF CVPR}, 2026.

\bibitem{openx2024}
Open X-Embodiment Collaboration, ``Open X-Embodiment: Robotic learning datasets and RT-X models,'' in \emph{Proc. IEEE ICRA}, 2024.

\bibitem{khazatsky2024droid}
A. Khazatsky \emph{et al.}, ``DROID: A large-scale in-the-wild robot manipulation dataset,'' in \emph{Proc. Robotics: Science and Systems}, 2024.

\bibitem{liu2023libero}
B. Liu, Y. Zhu, C. Gao, Y. Feng, Q. Liu, Y. Zhu, and P. Stone, ``LIBERO: Benchmarking knowledge transfer for lifelong robot learning,'' in \emph{Advances in Neural Information Processing Systems, Datasets and Benchmarks Track}, 2023.

\bibitem{gupta2023armbench}
A. Gupta \emph{et al.}, ``ARMBench: An object-centric benchmark dataset for robotic manipulation,'' in \emph{Proc. IEEE ICRA}, 2023.

\bibitem{tyrovolas2024ur3}
M. Tyrovolas, K. Aliev, D. Antonelli, and C. Stylios, ``UR3 CobotOps,'' UCI Machine Learning Repository, 2024, doi: 10.24432/C5J891.

\bibitem{cockroachvector}
Cockroach Labs, ``Vector indexes,'' 2026. [Online]. Available: \url{https://www.cockroachlabs.com/docs/stable/vector-indexes}. Accessed: Aug. 4, 2026.

\bibitem{awstitan}
Amazon Web Services, ``Amazon Titan Text Embeddings models,'' 2026. [Online]. Available: \url{https://docs.aws.amazon.com/bedrock/latest/userguide/titan-embedding-models.html}. Accessed: Aug. 4, 2026.
\end{thebibliography}
